\documentclass[runningheads]{llncs}
\usepackage{graphicx}
\usepackage{amsmath,amssymb} 

\usepackage{xspace}
\usepackage{comment}
\makeatletter
\DeclareRobustCommand\onedot{\futurelet\@let@token\@onedot}
\def\@onedot{\ifx\@let@token.\else.\null\fi\xspace}
 
\def\ie{i.e\onedot}

\def\etal{et al\onedot}
\usepackage{booktabs}
\usepackage[pagebackref=true,breaklinks=true,letterpaper=true,colorlinks,bookmarks=false]{hyperref}
\begin{document}
\pagestyle{headings}
\mainmatter
	\title{D\'{e}j\`{a} Vu:\\ \fontsize{11}{12}\selectfont{Motion Prediction in Static Images}}
	\titlerunning{Motion Prediction in Static Images}
	\authorrunning{Silvia L. Pintea, Jan C. van Gemert, Arnold W. M. Smeulders}
	\author{Silvia L. Pintea, Jan C. van Gemert, and Arnold W. M. Smeulders}
	\institute{Intelligent Systems Lab Amsterdam (ISLA), University of Amsterdam\\
		Science Park 904, 1098 HX, Amsterdam, The Netherlands}
\maketitle
\begin{abstract}
This paper proposes motion prediction in single still images by learning it from 
a set of videos. The building assumption is that similar motion is characterized 
by similar appearance. The proposed method learns local motion patterns given a 
specific appearance and adds the predicted motion in a number of applications. 
This work (i) introduces a novel method to predict motion from appearance in a 
single static image, (ii) to that end, extends of the Structured Random Forest 
with regression derived from first principles, and (iii) shows the value of adding 
motion predictions in different tasks such as: weak frame-proposals containing 
unexpected events, action recognition, motion saliency. Illustrative results indicate 
that motion prediction is not only feasible, but also provides valuable information 
for a number of applications. 
\end{abstract}
\section{Introduction}
\begin{figure}[b!]
	\centering
	\begin{tabular}{l*{5}{p{.192\textwidth}}}
	\includegraphics[width=0.192\textwidth]{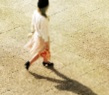} & 
	\includegraphics[width=0.192\textwidth]{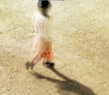} & 
	\includegraphics[width=0.192\textwidth]{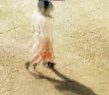} & 
	\includegraphics[width=0.192\textwidth]{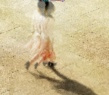} & 
	\includegraphics[width=0.192\textwidth]{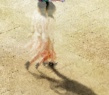} \\ 
	(a) & (b) & (b) & (b) & (b)\\ 
	\end{tabular}
	\caption{\small (a) Original still image. (b) Warps with different motion magnitude 
		steps --- obtained from predicted motion --- overlaid over original image.}
	\label{fig:idea}
\end{figure}
In human visual perception, expectation of what is going to happen next is essential 
for the on time interpretation of the scene and preparing for reaction when needed. 
The underlying idea in estimating motion patterns from a single image is illustrated 
by the walking person in figure~\ref{fig:idea}. This figure is obtained from the 
proposed method by warping the static image with the predicted motion at different 
magnitude steps. For a human observer it is obvious what to expect in figure~\ref{fig:idea}: 
motion in the legs and arms, while the torso moves to the right. From these clues 
we build our expectation. (That is the reason why the moonwalk by Michael Jackson 
is so salient as it refutes the expectation.) Closer inspection of figure~\ref{fig:idea} 
reveals that only the face, the legs and hands expose the expected motion direction, 
whereas the torso cannot reveal the difference to left or to the right. This indicates 
that motion is locally predictable. Therefore, for building motion expectation we 
start with local parts rather than the complete silhouette.

\begin{figure}[t!]
	\centering
	\includegraphics[width=0.9\textwidth]{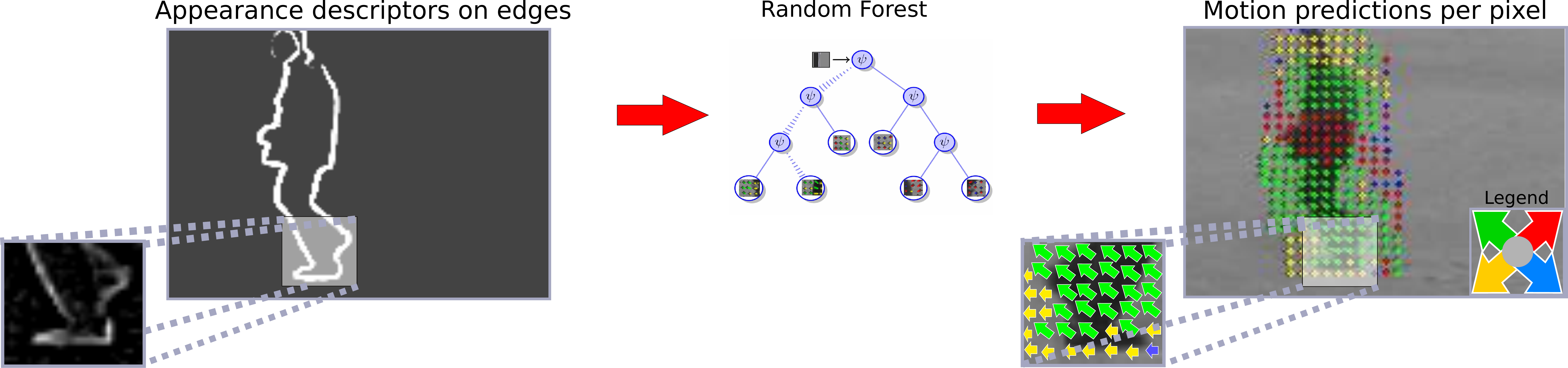}
	\caption{\small The Random Forest receives as input pairs of appearance-motion 
		patches and learns the correlation from the part-based appearance 
		to its corresponding motion.}
	\label{fig:motionPrediction}
\end{figure}

The paper shows that motion prediction in static images is possible by learning 
it from videos and transferring this knowledge to unseen static images. The proposed 
method does not use any localized object-class labels or frame-level annotations; 
it only requires suitable training videos. We consider a few applications of 
predicted motion: weak proposals of frames containing unexpected events, action 
recognition, motion saliency. This work has three main contributions: (i) a method 
for learning to predict motion in single static images from a training set of videos; 
(ii) to that end, extension of Structured Random Forests (SRF) with regression; 
(iii) a set of proposed possible applications for the motion prediction in 
single static images. Figure~\ref{fig:motionPrediction} illustrates the submitted approach.
\section{Related Work}
Adding motion to still images is also considered in SIFT flow~\cite{siftflow}, 
albeit with a different goal in mind --- global image aligning for scene matching. 
As opposed to this work, in~\cite{siftflow} two images are first aligned by matching local 
information and, subsequently, the motion from the training frame is transferred 
to the test frame. Other work visualizes motion in a static image by locally 
blurring along a vector field~\cite{cabralCG93imagingVectorFields} or creates a 
motion sensation illusion by varying oriented image filters~\cite{freemanCG91motionWOmovement}.
In~\cite{dai2008motion} the authors learn affine motion models from the blurring 
information in the $\alpha$-channel, while~\cite{cifuentes2012motion} proposes a 
grouping of point trajectories based on different types of estimated camera motion.
The authors of~\cite{kitani2012activity} focus on trajectory prediction by 
using scene information. Unlike these works, the proposed method predicts 
local motion by learning it from videos.

Local methods in video-based action recognition only depend on a pair of consecutive 
frames to compute the temporal derivative~\cite{ivo,klaserBMVC08hog3d} or optical 
flow~\cite{Laptev:CVPR08,Wang:BMVC09}. Because the temporal derivative lacks the 
motion direction and magnitude, this work focuses on predicting the more informative 
optical flow. Next to optical flow we also consider representing the motion as 
flow derivatives. These have been used in~\cite{dalal2006human} for MBH (Motion 
Boundary Histograms) computation.

Cross-modal approaches use static images to recognize actions in videos~\cite{opposite} 
and appearance variations in videos to predict and localize objects in static images
~\cite{tubes}. Similarly, we propose a cross-modal approach: learning from videos 
and applying the learned model to static images. 

Structured learning is suitable for this approach because motion is spatially 
correlated. Therefore, it is more appropriate to consider motion-patches, rather 
than look at pixel-wise motion vectors. Several structured learning approaches are 
available, such as CRF (Conditional Random Field)~\cite{laffertyICML01crf}, 
Structured SVM (Support Vector Machines)~\cite{tsochantaridis2006large} or Structured 
Random Forests~\cite{kont}. In this paper we use an SRF (Structured Random Forest) 
because it has innate feature selection and allows for easy parallelization.

In~\cite{kont}, the authors use SRF for semantic segmentation. Here, each 
feature-patch has a corresponding patch of semantic labels instead of a single pixel 
label. In contrast to Kontschieder \etal~\cite{kont}, we predict motion in static 
images. Thus, we work with continuous data (regression) where the labels are 
patches of measured motion vectors, not discrete classes. Apart from solving different 
problems: motion prediction versus image segmentation, we also perform different 
learning tasks: regression (continuous motion vectors) versus classification 
(pixel-level class labels). The more recent work of Doll{\'a}r \etal
~\cite{dollar2013structured} proposes the use of SRF for edge detection. 
Rather than estimating joint probabilities for the edge-pixels --- which would be 
prohibitively expensive --- they map their structured space to a discrete 
unidimensional space on which they evaluate the goodness of each split. Contrary 
to~\cite{dollar2013structured}, in this work the outputs are continuous motion 
vectors, thus mapping from patches of continuous vectors to a discrete space is 
not straightforwardly done and not without discarding useful information. 
\section{Motion Prediction --- Formalism}
Closer inspection of figure~\ref{fig:motionPrediction} reveals that the motion 
magnitude and direction are correlated with the appearance and they are 
consistent in a given neighborhood, therefore the problem requires structured 
output rather than single pixel-wise predictions. Thus, to learn motion 
from local appearance, this method uses a structured learning approach 
--- SRF --- which is fine tuned for regression.

\begin{figure}[t!]
	\centering
	\begin{tabular}{l*{2}{p{.440\textwidth}}}
	\includegraphics[width=0.440\textwidth]{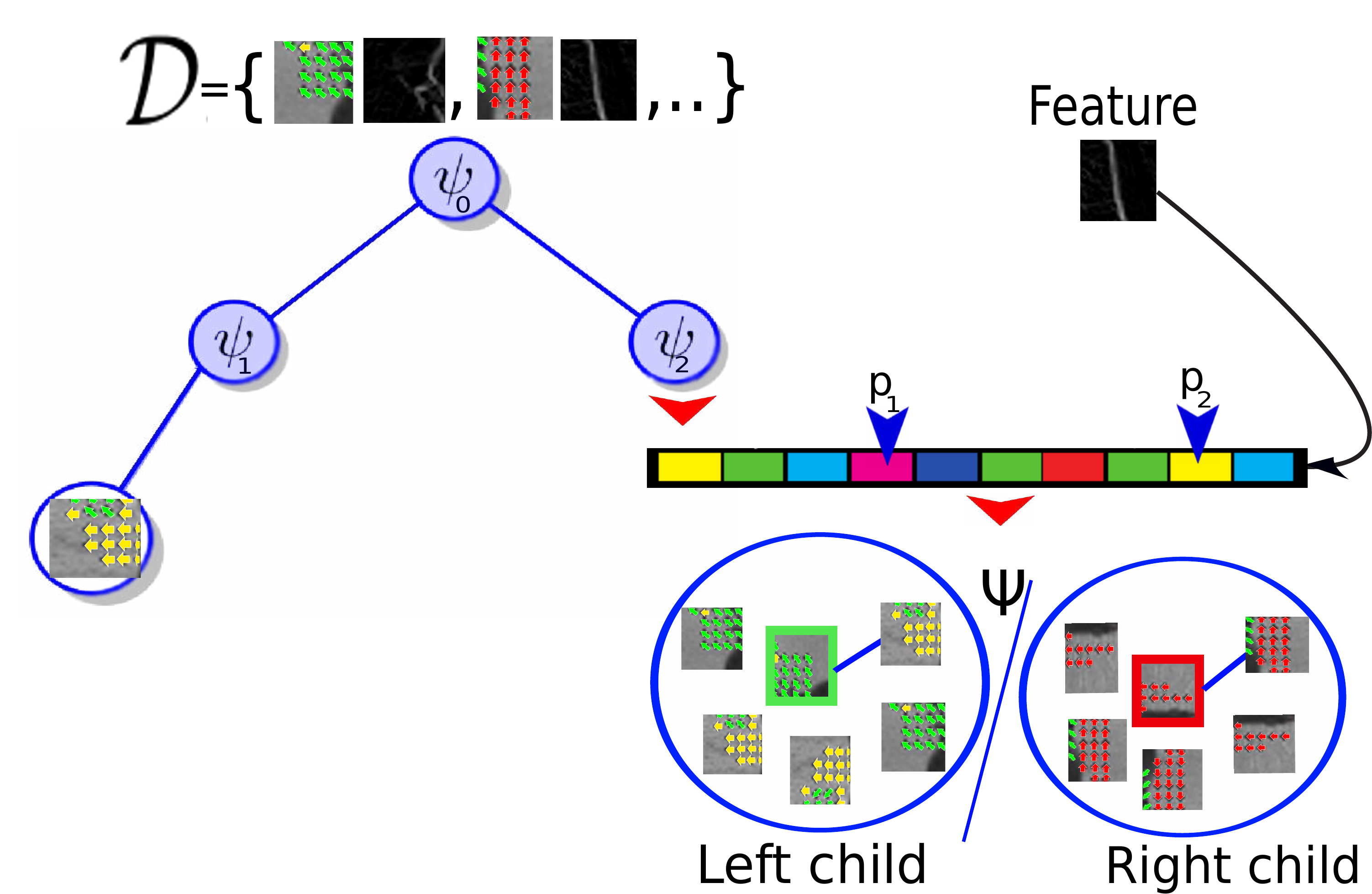}\hspace{30px} & 
	\includegraphics[width=0.440\textwidth]{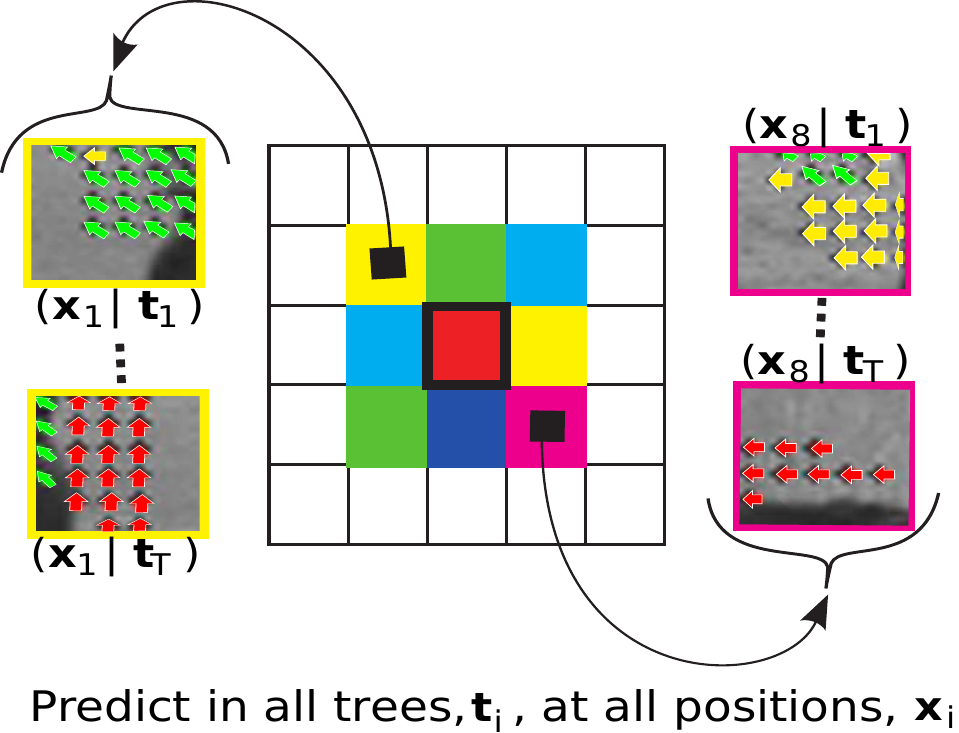}\\
	(a) & (b)\\ 
	\end{tabular}
	\caption{\small (a) Random Forest splitting: pixel-wise variance over the 
		continuous motion vectors is estimated for all the training 
		samples and all the splits. (b) Motion prediction in neighborhood: 
		the neighbors of the current patch contribute to the final prediction 
		at the current location due to overlap.}
	\label{fig:rftrain}
\end{figure}

A random forest is composed out of a number of trees. Each tree receives as input 
a set of training patches, $\mathcal{D}$, and their associated continuous motion 
patches --- spatial derivatives of optical flow or optical flow, as in 
figure~\ref{fig:rftrain}.a.\\[2px]
\textbf{Node splitting.} For each node in the tree, a number of splits $\Psi$, are 
generated by sampling: two random dimensions of the training features --- $\mathbf{F}$, 
denoted by $p_1$ and $p_2$, a random threshold $t$ and a split type that is randomly 
picked out of the following four split types:
\small 
\begin{alignat}{3}
	\Psi_1 &= \mathbf{F}(p_1) \ge t\hspace{30px} & \hspace{30px}\Psi_3 &= \mathbf{F}(p_1)-\mathbf{F}(p_2) \ge t\\
	\Psi_2 &= \mathbf{F}(p_1)+\mathbf{F}(p_2) \ge t\hspace{30px} & \hspace{30px}\Psi_4 &= \mid \mathbf{F}(p_1)-\mathbf{F}(p_2)\mid \ge t.
	\label{eq:splits}
\end{alignat}
\normalsize
Each generated split, $\Psi$, is evaluated at every training sample in the set 
$\mathcal{D}$. This decides if the sample goes to the left child, containing the 
values larger than $t$ or to the right child with values lower than $t$.\\[2px]
\textbf{Split Optimization.} At each node, the best split --- $\Psi^*$ --- is 
retained. The quality of a split is often evaluated by an information gain 
criterion~\cite{breiman2001random,kont}. Alternatively,~\cite{fanellofilter} 
optimizes the splits by minimizing a problem-specific energy formulation. 
Unlike in the common RF problems, here the predictions are continuous optical 
flow\slash flow derivative. This entails an SRF regression --- which is based 
on a motion similarity measure. 

In~\cite{trevor2001elements}, continuous predictions in the RF are approached 
by estimating the best split as the one corresponding to the smallest squared 
distance to the mean in each node. Despite its simplistic nature, the use of 
variance for regression in RF is also indicated as effective in practice 
in~\cite{criminisi2012decision,dollar2013structured}. In this case, the 
predictions are motion patches rather than single labels, which requires a 
measure of diversity of patterns inside these patches. For this purpose, 
we use pixel-wise variance:
\begin{alignat}{1}
	\mathcal{V}_{\mathcal{S}_\Psi} &= \sum_{\mathbf{x}\in{\mathcal{S}_\Psi}} \frac{1}{\mid P\mid} \sum_{i\in P}\frac{\sum_j^D (x_i^j-\mu_i^j)^2}{\mid {\mathcal{S}_\Psi}\mid-1},
	\label{eq:variance}
\end{alignat}
where $\mathcal{S}_\Psi \in \{\mathcal{L}_\Psi,\mathcal{R}_\Psi\}$ is the 
left\slash right child node, $P$ is the set of pixels in a patch --- 
$\mathbf{x}$, $D$ is the number of dimensions (\ie the 2 flow components) 
and $\mu_i^j$ is the mean motion at pixel $i$ and dimension $j$.

Consequently, the best split is the one characterized by minimum variance of 
patch patterns inside the two generated children:
\begin{alignat}{1}
	\Psi^* &= \text{argmin}_\Psi \frac{\mathcal{V}_{\mathcal{L}_\Psi} \mid \mathcal{L}_\Psi \mid + \mathcal{V}_{\mathcal{R}_\Psi} \mid \mathcal{R}_\Psi \mid }{\mid \mathcal{L}_\Psi \mid + \mid \mathcal{R}_\Psi \mid},
	\label{eq:split}
\end{alignat}
where $\mathcal{V}_{\mathcal{L}_\Psi\slash \mathcal{R}_\Psi}$ are the variances 
of the two child nodes (eq.~\ref{eq:variance}). The weighting by the node 
sizes is regularly used in RF to encourage more balanced splits~\cite{kont}.\\[2px] 
\textbf{Edge features.} The features used are patch-based HOG descriptors extracted 
over the opponent color space~\cite{van2010evaluating}. Given that motion can only 
be perceived at textured edge-patches (the aperture problem), we make the choice of 
extracting training patches for the SRF along the edges. This reduces the number of 
samples by retaining the relevant ones in terms of motion perception.\\[2px]
\textbf{Worst-first tree growing.} For stopping the tree growing and deciding to 
create a leaf, it is customary to threshold the variance 
(defined in eq.~\ref{eq:variance})~\cite{breiman2001random,criminisi2012decision,kont}. 
Rather than deciding to stop only based on variance thresholds, which can be 
cumbersome for different classes: \ie classes with larger motion magnitudes 
such as \emph{running} or \emph{jogging} have inherently higher motion variance 
than classes such as \emph{boxing}, we can impose a limit on the number of 
leaves we want in each tree. 

\cite{shotton2013decision} proposes the use of directed graphs for decision making. 
These graphs are iteratively optimized by alternating between finding the best 
splitting function and finding the best child assignment to parent nodes.     
Unlike here, the proposed ``worst-first" tree growing changes the order in which 
the nodes are visited, but not the topology of the trees. At each timestep 
the worst node --- with the highest variance as defined in equation~\ref{eq:variance} 
--- is chosen to be split next. 
Finally, when the number of terminal nodes reaches the number of desired leaves, 
the splitting process ends and all terminal nodes are transformed into leaves. 

By following this procedure we ensure that the more diverse nodes are split first, 
thus allowing for a fairly balanced tree at stopping time --- when the desired number 
of leaves is reached. Given that the data is more evenly split, this can be seen 
as a measure against overfitting as well as a speedup.\\[2px] 
\textbf{Leaf patch.} A leaf is created when all the patches in one node 
have a similar pattern. To enforce this, one can threshold the variance measure 
defined in equation~\ref{eq:variance} or additionally, as discussed above, grow 
the trees in a worst-first manner and select a desired number of leaves. 
At the point when a leaf node is created, it contains a set of patches, that 
are assumed to have a uniform appearance. Thus, we define the leaf prediction 
as the average over the patches in the node. This is a common approach for 
regression RF~\cite{criminisi2012decision,trevor2001elements} and, despite 
its simplicity, proves effective in practice.\\[2px]
\textbf{Motion Prediction.} Given an input test appearance patch, the goal is to 
obtain a motion prediction, $\textbf{x}^*$, from the trained SRF. For this, the 
method first evaluates at each edge-pixel in the image the most likely prediction 
over the trees of the SRF. Following the approach of~\cite{criminisi2012decision,trevor2001elements}
we define the prediction over trees as the average patch: 
$\mathbf{x}^* = \frac{1}{\mid \mathcal{T} \mid} \sum_{\mathbf{x}\in\mathcal{T}}\mathbf{x}$,
where $\mathcal{T}$ is the set of tree predictions at the current position. 
At last, because the patches overlap as shown in figure~\ref{fig:rftrain}.b, 
the final prediction at each pixel position is the average over all overlapping 
predictions at that position.
\section{Motion Prediction --- Learning and Features}
\begin{figure}[t!]
	\centering
	\begin{tabular}{l*{4}{p{.22\textwidth}}}
	\includegraphics[width=0.22\textwidth]{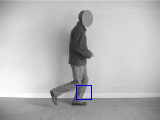} & 
	\includegraphics[width=0.22\textwidth]{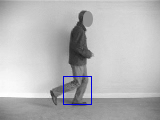}\hspace{15px} &
	\includegraphics[width=0.22\textwidth]{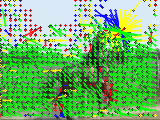} & 
	\includegraphics[width=0.22\textwidth]{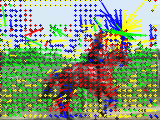}\\
	(a) Patch $11\times 11$ & (b) Patch $32\times 32$ & (c) Original flow & (d) Corrected flow\\ 
	\end{tabular}
	\caption{\small (a) \& (b) Patch size selection: small patch sizes fail 
		to capture motion direction, while large ones are more error prone. 
		(c) \& (d) Camera motion correction --- section~\ref{sec:appl}.3: 
		(c) original Farneb\"ack flow estimation, (d) affine corrected 
		flow estimation.}
	\label{fig:patch}
\end{figure}

\textbf{SRF patch size selection.} All experiments use the same patch size for 
features and for motion. Figures~\ref{fig:patch}.a and~\ref{fig:patch}.b show 
that a small patch size would fail to provide an indication for the expected 
motion direction. On the other hand, large patch sizes would be prone to 
prediction mistakes since full-body poses are characterized by larger motion 
diversity, thus, harder to learn than local patches. We have experimented 
with different patch sizes, and found a patch size approximately 5 times smaller 
than the maximum image size to be reasonable.\\[2px]
\textbf{Training features and labels.} The model learns motion from static 
appearance features at Canny edges to avoid the aperture problem. Motion is 
represented by dense optical flow, and optical flow derivatives respectively, 
measured at every pixel in the patch. For flow estimation we use the \emph{OpenCV} 
library\footnote{http://opencv.org}. One opponent-HOG descriptor is extracted 
over each training patch. Each HOG computation uses 2$\times$2 spatial bins, 
9 orientations and has 3 color-channels.\\[2px]
\textbf{SRF parameter settings.} The SRF uses 50 iterations per node during 
training and another 10 for finding the optimal threshold (eq.~\ref{eq:splits}). 
Given that the patch-features have 108 dimensions, the number of iterations --- 
usually set to the square root of the number of feature dimensions --- is sufficient. 
All trees are grown in the ``worst-first" manner until a number of 1,000 leaves 
is reached, additionally leaves are created when the variance (eq.~\ref{eq:variance}) 
goes below a 0.1 threshold. Each tree in the SRF is trained on 20 random pairs of 
sequential frames. 
\section{Evaluating Motion Prediction}
\label{sec:exp1}
This section focuses on evaluating the SRF motion predictions with respect to 
measured and ground truth video motion. The goal of this experiment is to test 
the ability to learn motion from appearance. For evaluation, we use both 
KTH~\cite{schuldtICPR04KTHset} and Sintel~\cite{Butler:ECCV:2012} datasets. 
The choice for the simplistic KTH dataset is motivated by its laboratory setting 
providing limited appearance variations and reliable optical flow measurements 
--- limited to no camera motion.\\[2px]
\begin{figure*}[t!]
	\centering
	\begin{tabular}{l*{5}{p{.185\textwidth}}}
		\includegraphics[width=0.185\textwidth]{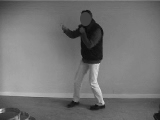} &
		\includegraphics[width=0.185\textwidth]{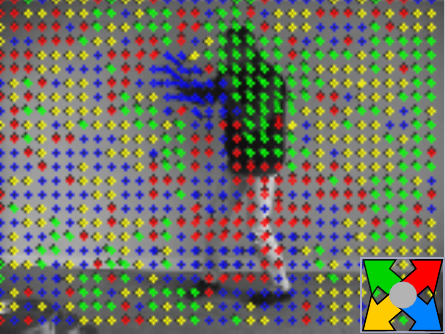} &
		\includegraphics[width=0.185\textwidth]{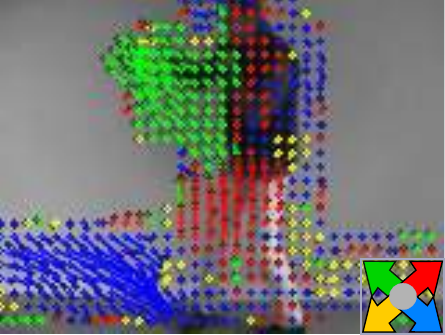} &
		\includegraphics[width=0.185\textwidth]{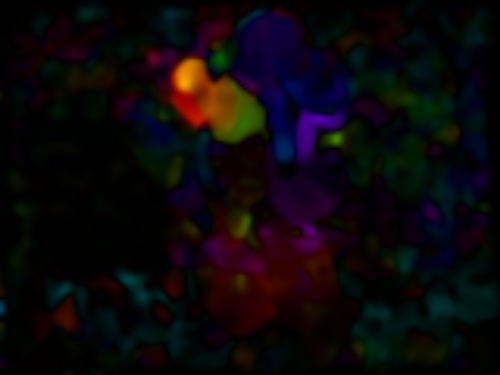} &
		\includegraphics[width=0.185\textwidth]{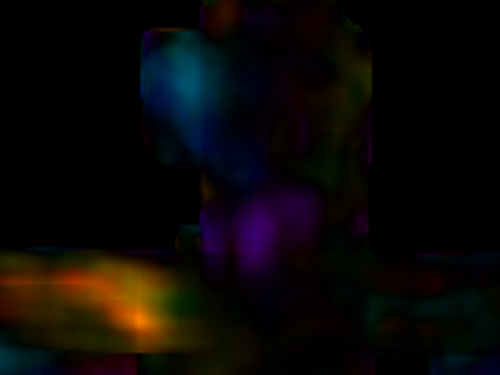}\\

		\includegraphics[width=0.185\textwidth]{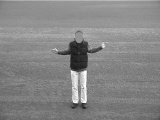} &
		\includegraphics[width=0.185\textwidth]{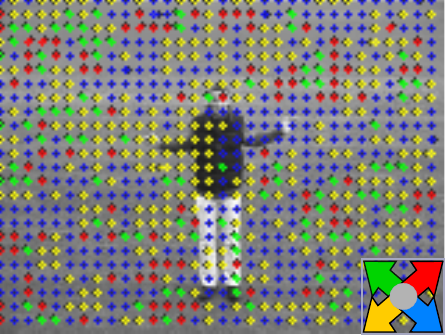} &
		\includegraphics[width=0.185\textwidth]{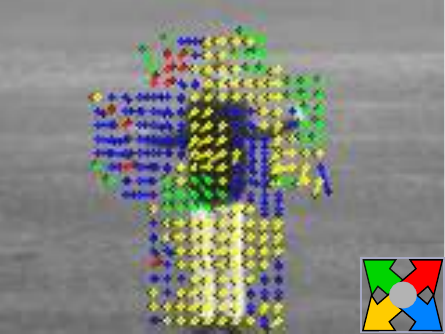} &
		\includegraphics[width=0.185\textwidth]{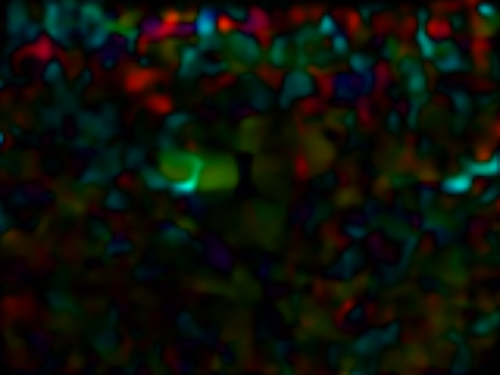} &
		\includegraphics[width=0.185\textwidth]{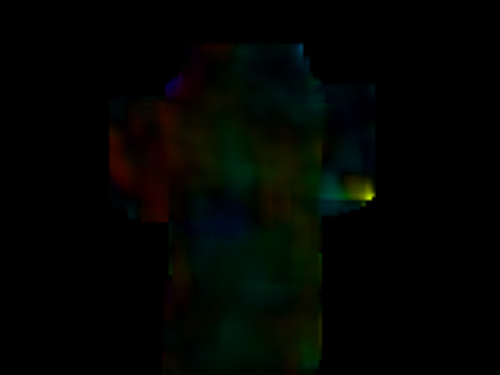}\\

		\includegraphics[width=0.185\textwidth]{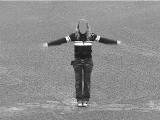} &
		\includegraphics[width=0.185\textwidth]{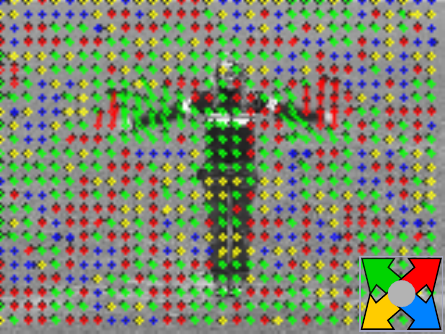} &
		\includegraphics[width=0.185\textwidth]{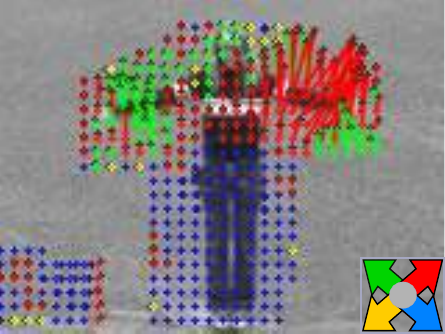} &
		\includegraphics[width=0.185\textwidth]{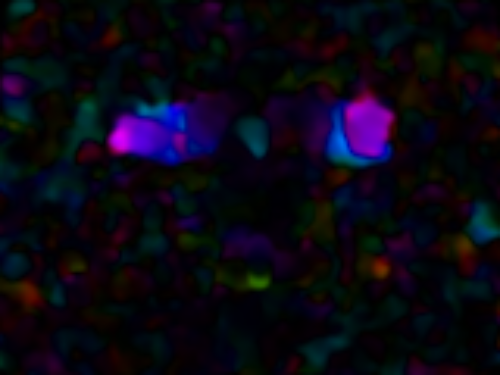} &
		\includegraphics[width=0.185\textwidth]{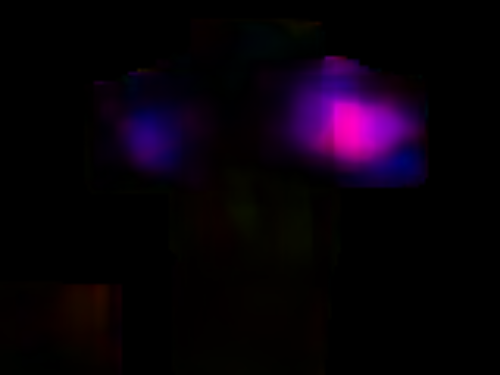}\\

		\includegraphics[width=0.185\textwidth]{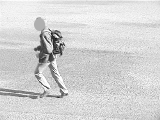} &
		\includegraphics[width=0.185\textwidth]{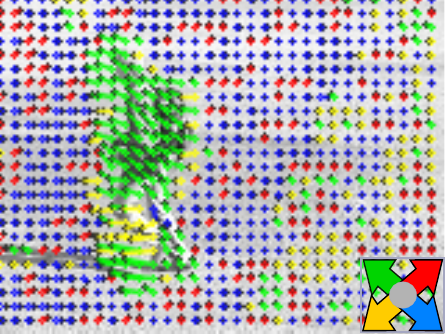} &
		\includegraphics[width=0.185\textwidth]{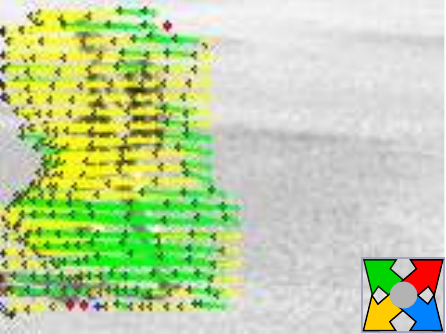} &
		\includegraphics[width=0.185\textwidth]{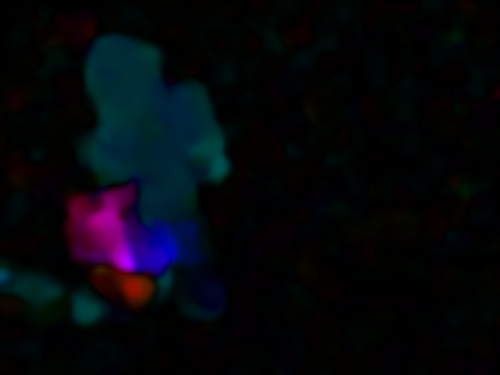} &
		\includegraphics[width=0.185\textwidth]{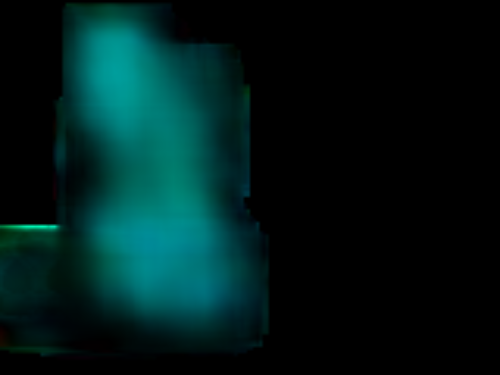}\\

		\includegraphics[width=0.185\textwidth]{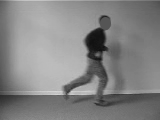} &
		\includegraphics[width=0.185\textwidth]{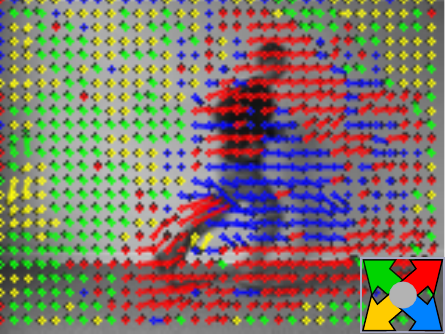} &
		\includegraphics[width=0.185\textwidth]{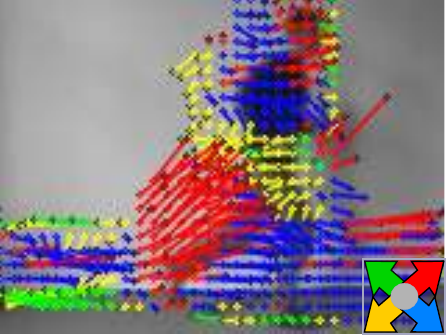} &
		\includegraphics[width=0.185\textwidth]{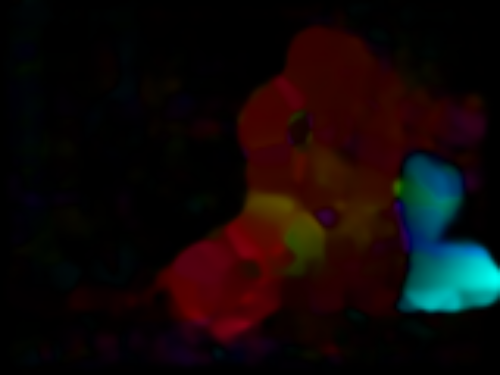} &
		\includegraphics[width=0.185\textwidth]{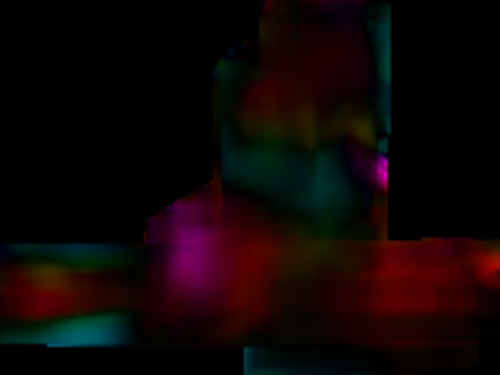}\\

		\includegraphics[width=0.185\textwidth]{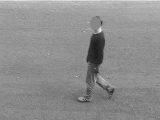} &
		\includegraphics[width=0.185\textwidth]{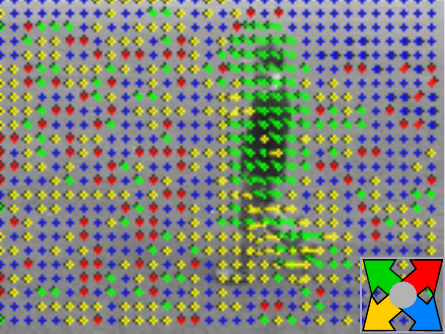} &
		\includegraphics[width=0.185\textwidth]{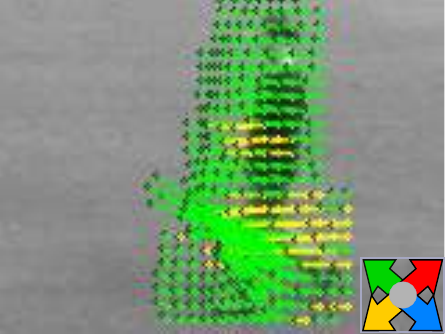} &
		\includegraphics[width=0.185\textwidth]{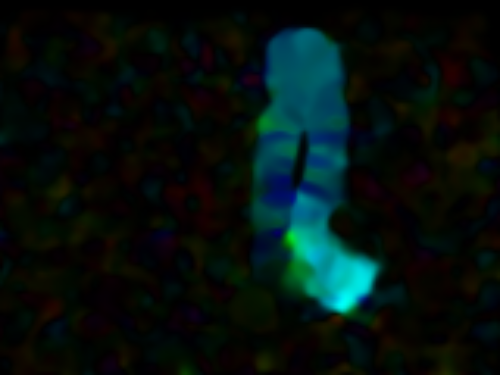} &
		\includegraphics[width=0.185\textwidth]{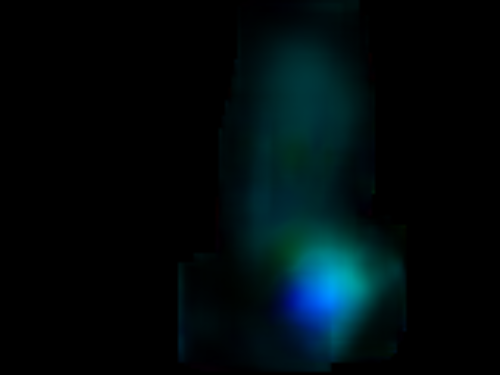}\\
		(a) & (b) & (c) & (d) & (e)\\ 
	\end{tabular}
	\caption{\small (a) Original image. (b) Measured optical flow vectors. 
		(c) Predicted optical flow vectors. (d) Measured flow map. (e) 
		Flow map of the predictions (colors as in~\cite{max1994visualizing}).} 
	\label{fig:resultsMotion}
\end{figure*}
\textbf{Setup.} We use the KTH dataset where the data is split in the standard 
manner~\cite{schuldtICPR04KTHset}. From each video we retain only a set of 20 
sequential frames. 
We compare the motion predictions against measured flow on the complete data --- 
trainval and test. For each one of the 6 classes we train an SRF regressor 
containing 11 trees. We train SRFs on the training set and use them to predict 
on the validation set and vice versa. Finally, we merge the SRFs of the two sets 
and use them for predicting on the test. 
Given that we extract motion patches along the Canny edges, we evaluate the 
motion predictions at the edge points. We use the standard optical flow error 
estimation --- EPE (End-Point-Error) --- which computes the Euclidian distance 
between the end point of the measured OF and the predicted one. We also 
evaluate the direction of predicted flow by computing the cosine similarity 
between the prediction and the measurement: 
$\frac{x_1 x_2 + y_1 y_2}{\sqrt{x_1^2 + y_1^2} \sqrt{x_2^2+y_2^2}}$. The orientation 
of the flow is estimated as the angle between the prediction and the measured 
OF on the half-circle: $\frac{\mid x_1 x_2 + y_1 y_2\mid}{\sqrt{x_1^2 + y_1^2} \sqrt{x_2^2+y_2^2}}$.\\[2px]
\begin{table}[t!]
	\centering
	\small{\begin{tabular}{l*{5}{p{.2\textwidth}}}\toprule
		 & Edge EPE & 0-Edge EPE & Edge direction & Edge orientation \\ \cmidrule(r){2-3}\cmidrule(l){4-5}
	Boxing & 1.23 px & \textbf{1.21} px & 3.86 \% & 66.60 \% \\ 
	Clapping & 1.13 px & \textbf{1.08} px & 1.18 \% & 65.80 \% \\ 
	Waving & 1.65 px & \textbf{1.59} px & -0.49 \% & 67.52 \% \\ 
	Jogging & \textbf{4.51} px & 4.61 px & 14.39 \% & 76.52 \% \\ 
	Running & \textbf{8.41} px & 8.62 px & 26.68 \% & 77.02 \% \\ 
	Walking & \textbf{3.90} px & 3.92 px & 10.80 \% & 72.17 \% \\ \cmidrule(r){2-3}\cmidrule(l){4-5}
	Avg. & \textbf{3.47} px & 3.50 px & 9.40 \% & 70.94 \% \\ 
	\bottomrule\\
	\end{tabular}}
	\caption{\small Predicted EPE (End-Point-Error) at the Canny edges compared 
		to the EPE of the zero prediction. Cosine similarity at edge-points 
		for direction estimation. Cosine similarity at edge-points using the 
		angle on the half-circle for orientation estimation. The direction 
		and orientation should be as close as possible to 100\%.}
	\label{tab:epe}
\end{table}
\textbf{Evaluation.} Figure~\ref{fig:resultsMotion} depicts a few examples of 
measured motion comparative to predicted motion and their corresponding flow maps. 
Noteworthy here is that the predicted motion is realistic and in agreement with 
the expectation of the observer. Moreover, the flow maps for the measured flow 
are similar to the flow maps of the predicted motion.
Table~\ref{tab:epe} shows that for \emph{jogging}, \emph{running} and \emph{walking} 
the EPEs are lower than the 0-prediction and also on average the predicted motion is 
better than the 0-EPE. While looking at the direction estimation, one can see that 
the first 3 classes are quite often wrong --- due to the characteristic 
bidirectional motion. On the other hand, the orientation of the flow is considerably 
better (closer to 100\%) especially for the last 3 classes that involve larger 
motion magnitudes --- \emph{jogging}, \emph{running} and \emph{walking}.\\[2px]
\begin{table}[b!]
	\centering
	\small{\begin{tabular}{l*{5}{p{.2\textwidth}}}\toprule
		 & Edge EPE & 0-Edge EPE & Edge direction & Edge orientation \\ \cmidrule(r){2-3}\cmidrule(l){4-4}\cmidrule(l){5-5}
	Horn-Schunck & 4.25 px & 4.21 px & 02.72 \% & 65.69 \%\\ 
	Lucas-Kanade & 3.18 px & 3.22 px & 08.30 \% & 70.03 \%\\ 
	Farneb\"ack & 3.47 px & 3.50 px & 09.40 \% & \textbf{70.94} \%\\ 
	Simple Flow &\textbf{0.91} px & \textbf{1.01} px & \textbf{18.30} \% & 70.08 \%\\ \bottomrule\\
	\end{tabular}}
	\caption{\small Average scores over classes at the Canny edges compared to 
		the scores of the zero prediction when the training of the SRFs 
		uses different flow algorithms: Farneb\"ack, Horn-Schunck, 
		Lucas-Kanade, Simple Flow.}
	\label{tab:epeFlow}
\end{table}
\textbf{Impact of flow algorithm.} Table~\ref{tab:epeFlow} shows the average scores 
over the classes when the training of the SRFs is based on different existing flow 
algorithms. Farneb\"ack and Simple Flow are characterized by smaller errors in both 
flow magnitude as well as flow direction and orientation --- Simple Flow has a 10\% 
relative gain over the 0-baseline in magnitude while Farneb\"ack gains a 3\% over 
the 0-baseline. Yet the direction of the prediction is more often correct when 
training on Farneb\"ack flow. 

We run an additional experiment on the Sintel dataset~\cite{Butler:ECCV:2012} for 
which ground truth flow is provided. We have retained 10 frames out each training 
video for learning the SRF and used the rest for testing. We have trained only one 
SRF containing 11 trees where each tree was trained on maximum 20 randomly sampled 
frame pairs. We have compared the predictions of the SRF when using the ground truth 
flow with the measured flow for both Simple Flow and Farneb\"ack. The scores 
are computed with respect to the ground truth flow. Table~\ref{tab:sintel} displays 
the results. Predicted flow outperforms Farneb\"ack measurements in terms of edge 
EPE, while the direction of the predicted flow is often correct.\\[2px]
\begin{table}[t!]
	\centering
	\small{\begin{tabular}{l*{4}{p{.25\textwidth}}}\toprule
		 & SRF prediction & Simple Flow & Farneb\"ack\\ \cmidrule(l){2-4}
	Edge EPE & 11.46 px & \textbf{09.26} px & 14.42 px \\ 
	Edge direction & 38.64 \% & 71.18 \% & \textbf{71.26} \%\\ 
	Edge orientation & 73.45 \% & 80.88 \% & \textbf{84.63} \%\\ \bottomrule\\
	\end{tabular}}
	\caption{\small Average scores over classes on Sintel data estimated at 
		Canny edges. The measured Farneb\"ack and Simple Flow are compared 
		to the SRF predictions.}
	\label{tab:sintel}
\end{table}
\textbf{Comparison with other learning methods.} Table~\ref{tab:otherLearn} 
compares the SRF motion predictions on the KTH data with motion predictions obtained 
by training a pixel-wise SVR (Support Vector Regressor) with linear kernel and 
pixel-wise least squares regression. We train a separate regressor for the x and y 
coordinates of the flow. The SRF obtains the smallest error --- 3\% relative gain 
over the 0 baseline while the other two methods perform worse than the 0-baseline. 
This experiment ascertains the need for a structured learning regression method.
\begin{table}[t!]
	\centering
	\small{\begin{tabular}{l*{5}{p{.2\textwidth}}}\toprule
		 & 0-prediction & SRF & SVR & Least squares\\ \cmidrule(l){2-5}
	Edge EPE & 3.50 px & \textbf{3.47} px & 5.02 px & 6.09 px\\ \bottomrule\\
	\end{tabular}}
	\caption{\small Average EPE over classes on KTH data estimated at Canny 
		edges when all methods are trained on Farneb\"ack flow.}
	\label{tab:otherLearn}
\end{table}
\section{Applications of Motion Prediction}
\label{sec:appl}
In this section, we bring forth a few possible uses of the predicted motion. One 
could consider other applications, as this is not a complete list of tasks that 
could benefit from motion prediction. \textbf{Application 1} gives an illustration 
of weakly detecting unexpected events in videos. \textbf{Application 2} evaluates 
how far off is the predicted motion from the measured motion in the context of 
action recognition. \textbf{Application 3} focuses on the gain of adding motion 
prediction to action recognition in still images, while \textbf{application 4} 
proposes a method for motion saliency in static images.
\subsection{Application 1 --- Illustration of finding unexpected events}
\begin{figure}[t!]
	\centering
	\begin{tabular}{c}
	\includegraphics[width=0.95\textwidth]{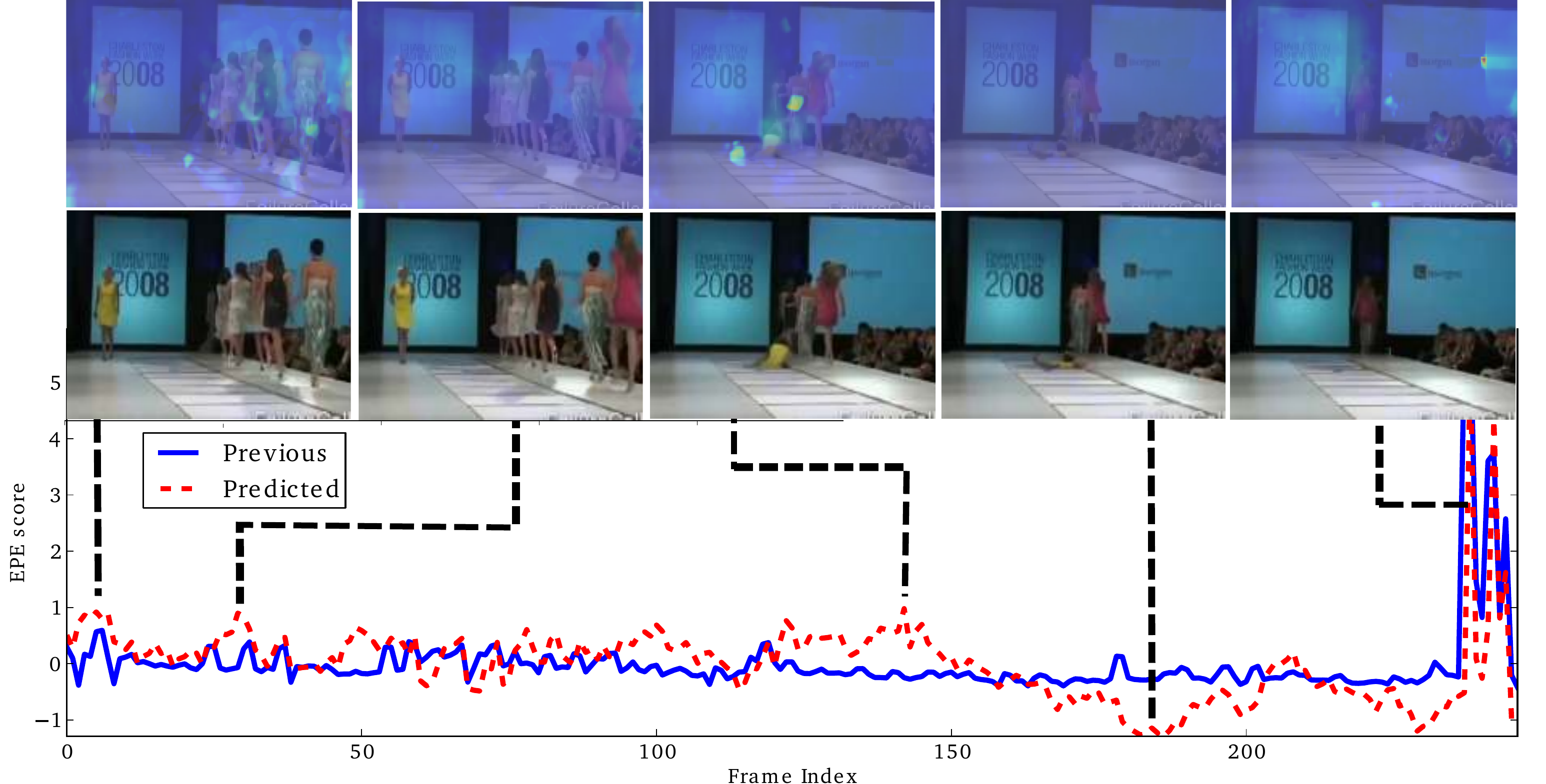}\\ 
	\end{tabular}
	\caption{\small EPE between measured flow at previous frame and 
		current frame in blue. EPE between measured flow at current 
		frame and predicted flow at current frame in red. The two 
		graphs are centered on 0 for illustration purpose. Images 
		containing unexpected motion together with their associated 
		EPE heatmaps are displayed on top.}
	\label{fig:unexpected}
\end{figure}
This application shows a possible use of the motion prediction --- finding unexpected 
events. We do not present this as an end goal of the proposed motion prediction 
method, but rather as an illustration of its usefulness. Given an SRF trained on 
a set of videos, one could use the SRF to obtain motion predictions at each frame 
in a given unseen video. If the EPE between the measured motion vectors and the 
predicted ones is large, it can be assumed that a motion that has not been seen 
before in the training data occurs at some point in the video.\\ 
\textbf{Setup.} For speed considerations, the SRF uses 3 trees. 
The training videos are queried from the TRECVid (TREC Video Retrieval Evaluation) 
\cite{trecvid,van2009episode} development set. The frames are resized to maximum 
300 px. We obtain motion predictions at each frame in the test video and compute 
the EPE score between the measured Farneb\"ack optical flow and the predicted flow. 
We retain the frames whose EPE scores are over one standard deviation from the mean. 
The same procedure is repeated for the baseline comparison, but this time by 
computing the EPE between the motion at the previous frame and the current frame.\\[2px]
\textbf{Evaluation.} Figure~\ref{fig:unexpected} displays the EPE error over the 
video frames, with respect to both previous frame and predicted motion. Frames 
containing interesting motion are displayed together with their corresponding 
EPE heatmaps. The video contains a catwalk show during which one person falls 
through the floor --- unexpected event. 
Noteworthy is that the predicted motion graph is reasonably similar to the one based 
on the errors to the previous frame, while better emphasizing the interesting 
moments --- when the person falls or is no longer visible in the frame --- 
through the graph peaks. 
\subsection{Application 2 --- Predicted vs. measured motion for AR}
Being able to predict motion can provide useful information in the 
action recognition context. The predictions obtained from appearance can be combined 
with appearance features in an action recognition pipeline. We present this application 
from a compelling theoretical perspective as we are not interested in absolute 
numbers but in gaining insight about the feasibility of motion prediction and its 
relative usefulness when compared to measured motion.\\[2px] 
\textbf{Setup.} We evaluate on the same KTH datatset. For action recognition, 
we follow the setting of Laptev \etal~\cite{Laptev:CVPR08}. We extract HOG 
features for appearance description, and HOF, MBH respectively for motion 
description at Canny edges. We also use a level-2 spatial pyramid for all 
descriptors~\cite{lazebnikCVPR06spatialPyr}. For each class we train a 
one-vs-all SVM classifier with HIK (Histogram Intersection Kernel) where 
the $C$ parameter is set by performing 5-fold crossvalidation on the 
subsampled trainval set. The obtained predictions from all 6 one-vs-all 
SVMs are ranked.\\[2px]
\begin{table*}[b!]
	\centering
	\small{\begin{tabular}{l*{8}{p{.115\textwidth}}}\toprule
	 & HOG & \multicolumn{4}{c}{Predicted Motion} & \multicolumn{2}{c}{Measured motion}\\ \cmidrule(r){3-6}\cmidrule(l){7-8} 
	 & & pHOF & HOG+ pHOF & pMBH & HOG+ pMBH & mHOF & mMBH\\ \cmidrule(r){3-6}\cmidrule(l){7-8} 
Boxing & 0.58 & 0.56 & \underline{0.58} & \underline{0.61} & \underline{\textbf{0.70}} & 0.75 & 0.78 \\ 
Clapping\hspace{10px} & 0.83 & 0.64 & \underline{0.89} & \underline{0.87} & \underline{\textbf{0.92}} & 0.89 & 0.89 \\ 
Waving & 0.97 & 0.89 & \underline{\textbf{0.97}} & 0.86 & 0.94 & 1.00 & 0.94 \\
Jogging & 0.72 & 0.61 & \underline{\textbf{0.78}} & 0.56 & 0.70 & 0.78 & 0.81 \\ 
Running & 0.83 & \underline{0.83} & \underline{\textbf{0.89}} & \underline{0.86} & \textbf{\underline{0.89}} & 0.81 & 0.86 \\ 
Walking & 0.97 & \underline{0.97} & \underline{\textbf{1.00}} & 0.80 & 0.91 & 0.89 & 0.97 \\ \cmidrule(r){3-6}\cmidrule(l){7-8}
Avg. & 0.82 & 0.75 & \underline{\textbf{0.85}} & 0.76 & \underline{0.84} & 0.85 & 0.88 \\ 
	\bottomrule\\
	\end{tabular}}
	\caption{\small Action Recognition accuracy on the KTH dataset for HOG only, 
		predicted\slash measured HOF --- pHOF\slash mHOF --- and predicted\slash 
		measured MBH --- pMBH\slash mMBH. The underlined text shows where 
		the prediction results are better than static (HOG) while the bold 
		shows the best.}
	\label{tab:kthAcc}
\end{table*}
\textbf{Evaluation.} Figure~\ref{fig:resultsMotion} shows a few examples of 
measured motion compared to predicted motion on KTH data, together with their 
corresponding flow maps. The accuracies for action recognition are displayed 
in table~\ref{tab:kthAcc}. The scores are lower than standarly reported in the 
literature due to the fact that we only use 20 frames per video. Here we compare 
the static features (HOG) with the combination of HOG and predicted motion 
(HOF and MBH). The measured motion exceeds the predicted motion in itself, 
but in combination with the appearance, the predicted motion can actually reach 
the scores of the measured motion. As expected, adding the motion information 
improves for categories that involve a larger amount of motion such as: 
\emph{running}, \emph{jogging} and \emph{walking} and less so for \emph{boxing}, 
\emph{handclapping} and \emph{handwaving}. It is worth noting that the combination 
of predicted HOF and HOG equals the measured HOF for this dataset. This proves 
that motion, even imperfect, brings useful information. 
\subsection{Application 3 --- Predicted motion for AR in Still Images}
While application 2 estimates how far apart are the predicted motion and 
the measured motion in the context of action recognition, here we predict 
motion over inherently static images which lack the video compression 
artifacts or motion blurring. The models are trained on realistic video-data 
and this application analyzes if the learned motion can, indeed, be transferred 
to still images. We measure the value of the static motion predictions in 
the context of image-based action recognition. Again, the goal here is to 
test the ability of predicting motion and its added value, and less so to 
improve over state-of-the-art.\\[2px] 
\begin{figure*}[t!]
	\centering
	\begin{tabular}{l*{4}{p{.25\textwidth}}}
		\includegraphics[width=0.25\textwidth]{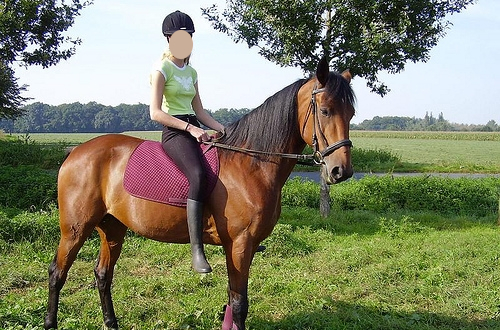} &
		\includegraphics[width=0.25\textwidth]{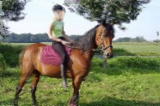} &
		\includegraphics[width=0.25\textwidth]{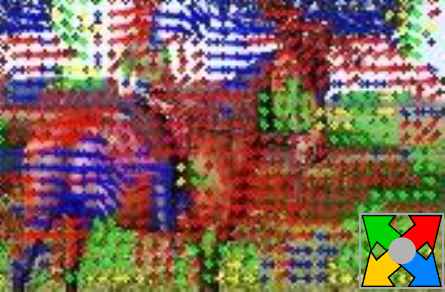} &
		\includegraphics[width=0.25\textwidth]{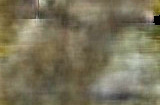}\\

		\includegraphics[width=0.25\textwidth]{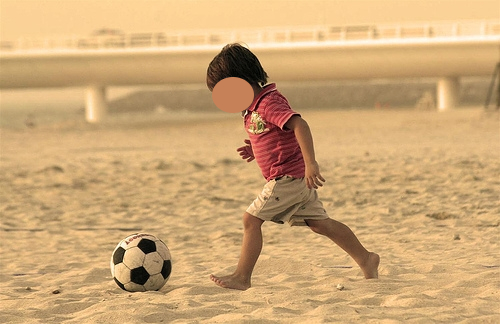} &
		\includegraphics[width=0.25\textwidth]{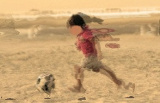} &
		\includegraphics[width=0.25\textwidth]{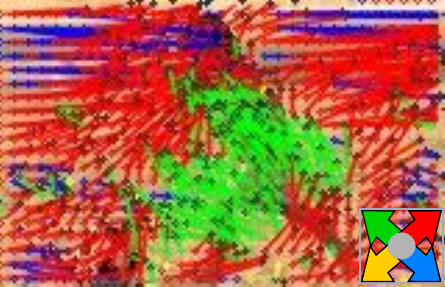} &
		\includegraphics[width=0.25\textwidth]{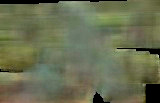} \\

		\includegraphics[width=0.25\textwidth]{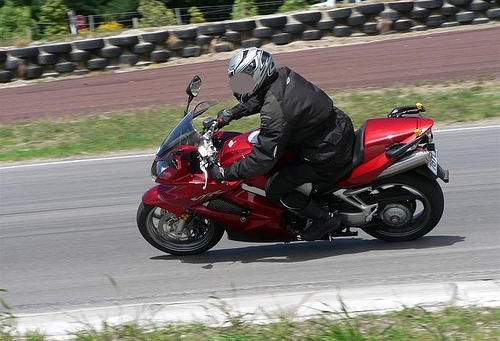} &
		\includegraphics[width=0.25\textwidth]{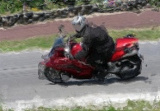} &
		\includegraphics[width=0.25\textwidth]{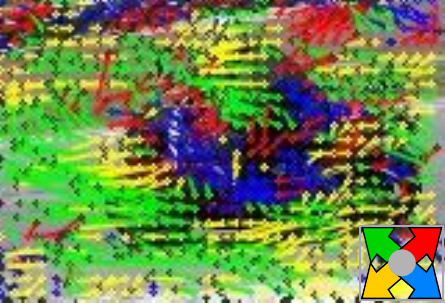} &
		\includegraphics[width=0.25\textwidth]{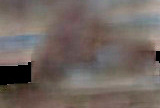} \\
		(a) & (b) & (c) & (d) \\ 
	\end{tabular}
	\caption{\small (a) Examples of images from the static Willow dataset
		~\cite{base}. (b) Warped image using the predicted motion, overlaid 
		over the original image. (c) Predicted flow vectors. (c) The training 
		appearance associated in the SRF with the predictions. More examples of 
		static images animated with predicted motion can be found at:
		\href{http://silvialaurapintea.github.io/dejavu.html}{http://silvialaurapintea.github.io/dejavu.html}.} 
	\label{fig:actions}
\end{figure*}
\textbf{Setup.} We apply the motion predictions on a static action recognition 
dataset --- Willow~\cite{base}. For each class we train a separate SRF on 
TRECVid data as in application 1. For each image we obtain seven flow\slash flow 
derivative predictions --- one per class, subsequently used for HOF\slash MBH 
descriptor extraction. In the action recognition part we use the same setting 
as in application 2, except that we extract descriptors densely over the images.\\[2px]
\textbf{Affine camera motion correction.} The videos used for training SRFs are
realistic videos characterized by large camera motion which drastically affects the 
Farneb\"ack measurements. To correct for camera motion we assume an affine motion 
model whose parameters we determine by first selecting a set of interest points in 
the two sequential frames. These interest points are matched between frames 
and the consistent matches are retained by employing RANSAC. From the kept 
point-matches we estimated the parameters of the affine model. Given the affine 
camera motion estimation, we then correct the second frame for this motion and 
subsequently perform flow estimation. Figures~\ref{fig:patch}.c and~\ref{fig:patch}.d 
show an example of original flow estimation and camera-motion corrected flow.\\[2px]
\textbf{Evaluation.} Figure~\ref{fig:actions} displays a few examples from the 
static dataset together with their predicted motion vectors, the appearance 
associated in the SRF with that motion and the static image warped with the 
predicted motion overlaid on top of the original image. 
Interesting to notice in figure~\ref{fig:actions} is that the SRF manages 
to distinguish the foreground motion from the background motion. We notice 
here, that unlike in the case of KTH predictions (figure~\ref{fig:resultsMotion}), 
there is predicted background flow, but this flow has at all times a different 
direction from the foreground motion. 

The results of Delaitre \etal~\cite{base} (0.63 MAP) exceed the proposed motion 
prediction results due to the more sophisticated model and more fine-tuned features. 
Provided that we would process our features --- HOG and HOF\slash MBH descriptors --- 
in the manner described in~\cite{base}, the relative improvement should remain. 
Despite the drawbacks of noisy motion estimations (\ie figures~\ref{fig:patch}.c 
and~\ref{fig:patch}.d), low video quality as well as large camera motion, the 
SRFs are capable of learning the motion patterns characteristic for each class. 
By adding predicted HOF to the static HOG descriptors we obtain a relative gain 
of 1\% in MAP -- from 50\% to 51\% --- and a 2\% relative gain in accuracy. 
\subsection{Application 4 --- Motion saliency}
Motion saliency is yet another possible use for the predicted motion. Objects that 
move differently from their background are salient and capture the viewer\rq s 
attention. Being able to predict motion in static images provides the advantage 
of finding pixels that can be distinguished from their surrounding pixels through 
their motion. Inspired by~\cite{GemertICMR08photo}, we propose descriptor pooling 
based on predicted flow. Rather than pooling image descriptors on spatial location 
only, here we also pool them on their predicted motion.\\[2px] 
\begin{figure}[t!]
	\centering
	\begin{tabular}{l*{3}{p{.3\textwidth}}}
	\includegraphics[width=0.3\textwidth]{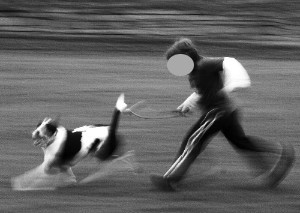} & 
	\includegraphics[width=0.3\textwidth]{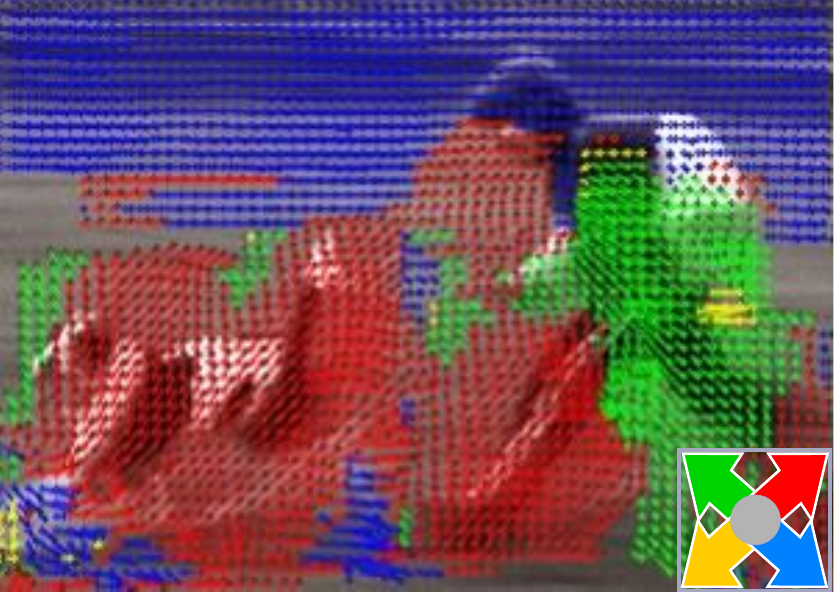} & 
	\includegraphics[width=0.3\textwidth]{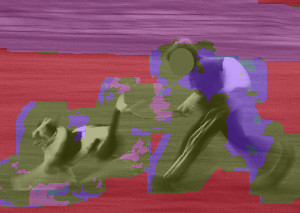}\\ 
	(a) & (b) & (c) \\ 
	\end{tabular}
	\caption{\small (a) Original still image. (b) Predicted flow vectors in the still 
		image. (c) Grouping descriptors based on their predicted motion --- 
		each color represents a pool in the flow-based spatial pyramid.}
	\label{fig:flowSP}
\end{figure}
\textbf{Setup.} This application uses the same experimental setup as 
application 3. We add to the spatial pyramid 10 more pools based on the predicted 
motion: 4 pools for the quadrants of the flow angles, 1 pool for the 0 prediction 
times 2 pools for flow magnitude larger\slash equal to 0.\\ 
\textbf{Evaluation.} Figure~\ref{fig:flowSP} displays an example of a static image 
together with its associated predicted motion vectors from the SRF and the corresponding 
motion grouping in the motion-based pooling framework. We notice that the background is 
grouped into a different pool than the foreground and also the pixels associated with 
the dog are grouped together while the ones corresponding to the boy are organized into 
a separate group.

By adding motion based pooling to static images, we obtain a 1\% relative improvement 
in both MAP and accuracy over the results in application 3. Thus, overall we have a 
relative gain on 2\% in MAP and 3\% in accuracy over static HOG features by adding 
motion --- predicted HOF features and flow-based pooling. Adding the flow-based 
pooling leads to improvements especially for the classes that involve a larger 
amount of motion: \emph{riding bike}, \emph{riding horse} and \emph{walking}. 
The outcome of this application ascertains that grouping pixels that have similar 
motion brings conclusive information. 
\section{Discussion}
The experimental evaluation tests the ability of the proposed method to learn 
motion given appearance features. One limitation of the method is 
data dependency --- the learned motion depends on the quality of the training data
and its similarity to the test data. This can be observed in application 3 
where motion is learned from complex realistic videos characterized by noise, motion 
blur and camera motion, yet the predictions are tested on high quality static images. 
Another drawback is the class dependency --- each action class is characterized by 
a certain direction and magnitude in the motion of the part (\ie arms, legs) and 
the SRFs learn class specific motion (figure~\ref{fig:resultsMotion}). 
Nonetheless, the considered applications show that learning motion from 
appearance is possible. Finally, the SRF code is made available online at: 
\href{http://silvialaurapintea.github.io/dejavu.html}{http://silvialaurapintea.github.io/dejavu.html}.
\section{Conclusions}
This paper proposes a method for motion prediction based on structured regression 
with RF. The undertaken task of performing structured regression in Random Forest 
is novel, as well as the problem of learning to predict motion in still images. 
We experimentally provide an answer to our first research question: can we learn 
to predict motion from appearance only? And we prove that this is possible provided 
proper training data and reliable optical flow estimates.
Furthermore, for illustrative purposes, we apply the proposed motion prediction 
method to a set of tasks and validate that the predicted, imperfect, motion adds 
novel and useful information over the static appearance features only, which answers 
our second research question. 
Finally, motion prediction can be employed in a multitude of topics such as: action 
anticipation or conflict detection. Another possible use for motion prediction is 
camera motion removal --- if the training videos are characterized by camera motion, 
the SRF is bound to learn the distinction between foreground and background motion 
(fig.~\ref{fig:actions},~\ref{fig:flowSP}).\\[7px]
\textbf{Acknowledgements.}\\
This research is supported by the Dutch national program COMMIT.
\bibliographystyle{splncs03}
\bibliography{main}
\end{document}